\documentclass{article}

 \usepackage[final]{neurips_2020}

\usepackage[utf8]{inputenc} 
\usepackage[T1]{fontenc} 
\usepackage{url} 
\usepackage{booktabs} 
\usepackage{amsfonts} 
\usepackage{nicefrac} 
\usepackage{microtype} 
\usepackage[dvipsnames]{xcolor}
\usepackage{wrapfig}
\usepackage{graphicx}
\usepackage{soul}
\graphicspath{{./figures/}}

\title{A modular framework for extreme weather generation}

\author{%
 Bianca Zadrozny \thanks{Authors listed alphabetically.}\\
 IBM Research \\
 \And
 Campbell D. Watson \footnotemark[1]\\
 IBM Research \\
  \And
 Daniela Szwarcman\footnotemark[1] \\
 IBM Research\\
 \And
 Daniel Civitarese\footnotemark[1] \\
 IBM Research \\
 \And
 Dario Oliveira\footnotemark[1]\\
 IBM Research \\
 \And
 Eduardo Rodrigues\footnotemark[1] \\
 IBM Research \\
 \And
 Jorge Guevara\footnotemark[1] \\
 IBM Research \\
}

\begin{document}

\maketitle

\begin{abstract}
Extreme weather events have an enormous impact on society and are expected to become more frequent and severe with climate change. In this context, resilience planning becomes crucial for risk mitigation and coping with these extreme events. Machine learning techniques can play a critical role in resilience planning through the generation of realistic extreme weather event scenarios that can be used to evaluate possible mitigation actions. This paper proposes a modular framework that relies on interchangeable components to produce extreme weather event scenarios. We discuss possible alternatives for each of the components and show initial results comparing two approaches on the task of generating precipitation scenarios.

\end{abstract}

\section{Introduction}


It is now well established that the frequency, duration, and intensity of different types of extreme weather events has been increasing as the climate system warms. For example, climate change leads to more evaporation that may exacerbate droughts and increase the frequency of heavy rainfall and snowfall events \cite{NAS2016}. These extreme weather events often result in extreme conditions or impacts, either by crossing a critical threshold in a social, ecological, of physical system, or by occurring simultaneously with other events \cite{Sene2012}. 

Sectors such as agriculture, water management, energy, and logistics traditionally rely on seasonal (1 to 12 months lead time) forecasts of climate conditions for planning their operations. Given the inherit uncertainty in seasonal climate forecasting, stochastic weather generators are often used to provide a set of plausible climatic scenarios, which are then fed into impact models for resilience planning and risk mitigation. A variety of weather generation techniques have been developed over the last decades, however they are often unable to generate realistic extreme weather scenarios, including severe rainfall, wind storms or droughts \cite{Verdin2018}.

Here we propose a modular framework for weather generation that explicitly takes extreme weather events into account. The modularity comes from the fact that the internal framework components are model-agnostic. However, we emphasize the use of state-of-the-art machine learning (ML) techniques because of the potential to facilitate communication between the internal components. We believe this is the case because ML models allow more flexibility than traditional statistical or physics-based models in the definition of what constitutes an extreme weather event, and because ML models can be easily conditioned or parameterized by variables of interest.
Furthermore, the framework allows breaking the extreme weather generation task into simpler subtasks where the application of ML techniques is more direct. In practice, because of its modularity, the framework instantiates different pipelines that are useful for research and experimentation.

\section{Stochastic Weather Generation}

Stochastic weather generators (WGs) are models designed to simulate realistic time series of atmospheric variables, such as precipitation and temperature \cite{Benoit2020, Peleg2017}.They can be used to provide a variety of plausible climatic scenarios which are required for risk assessment in impact studies \cite{Verdin2018}. In the area of hydrology, for example, WGs are used to generate precipitation time series that are necessary to estimate flood risk or evaluate the sensitivity of the hydrological regime to climate change \cite{Peleg2017}.

There are several WGs in the literature founded on a variety of methods \cite{Peleg2017}. Traditional WGs first model the precipitation occurrence to generate time series of wet and dry spells, and then the precipitation intensity for the wet days is parameterized using probability distributions \cite{Peleg2017, Verdin2018}. Other variables, such as temperature, are then generated and cross-correlated with the wet-dry sequence \cite{Peleg2017}. Researchers have also proposed nonparametric WGs, using the k-nearest neighbor resampling method, for example \cite{Verdin2018}.

Despite the significant progress in weather generation techniques, many still fail to represent extreme events, such as severe rainfalls or droughts \cite{Verdin2018}. Evin et al. \cite{Evin2018} try to address this problem by extending a traditional multisite WG to use a heavy-tailed distribution to model the precipitation values. Other ideas have also been proposed, such as modeling the precipitation occurrence with more than the two typical dry and wet states \cite{Flecher2010} or perturbing the nonparametric WGs \cite{Lee2012}.

Another critical point is that, in the context of climate change impact analysis, the WGs must generate scenarios that are consistent with future climate information \cite{Verdin2018}. One possible approach is to condition the WG on seasonal predictions: Verdin et al. \cite{Verdin2018} include domain-averaged seasonal total precipitation and mean, maximum and minimum temperatures as covariates in their parametric WG. The authors in \cite{Benoit2020} also use meteorological covariates to condition their rain type series generation. Peleg et al. \cite{Peleg2019} introduce a different idea: they use modified climate statistics inferred from predictions of Global Circulation Models (GCMs) or Regional Climate Models (RCMs) to reparametrize their WG.

Impact studies demand realistic scenario generation, especially for extreme events consistent with historical data and future projections. In this sense, we aim to explore state-of-the-art deep generative models, such as variational autoencoders and generative adversarial networks (GAN), to synthesize realistic weather sequences. These ML models learn straight from data, thus relaxing many of the constraints commonly observed in physical or even statistical models. Besides serving as surrogates of complex phenomena, they disregard intricate mandatory parameters for controlling the events themselves -- unless strictly modeled to do so -- and can constitute a powerful weather generation tool using simple control variables.

Some researchers have presented initial results using GANs \cite{gan1, gan2} to generate weather sequences. However, there is still room to investigate these models in their ability to represent extreme events and also explore the use of seasonal forecasts as control variables.

\section{Extreme weather generation framework}

\begin{figure}[h]
\centering
\includegraphics[width=\textwidth]{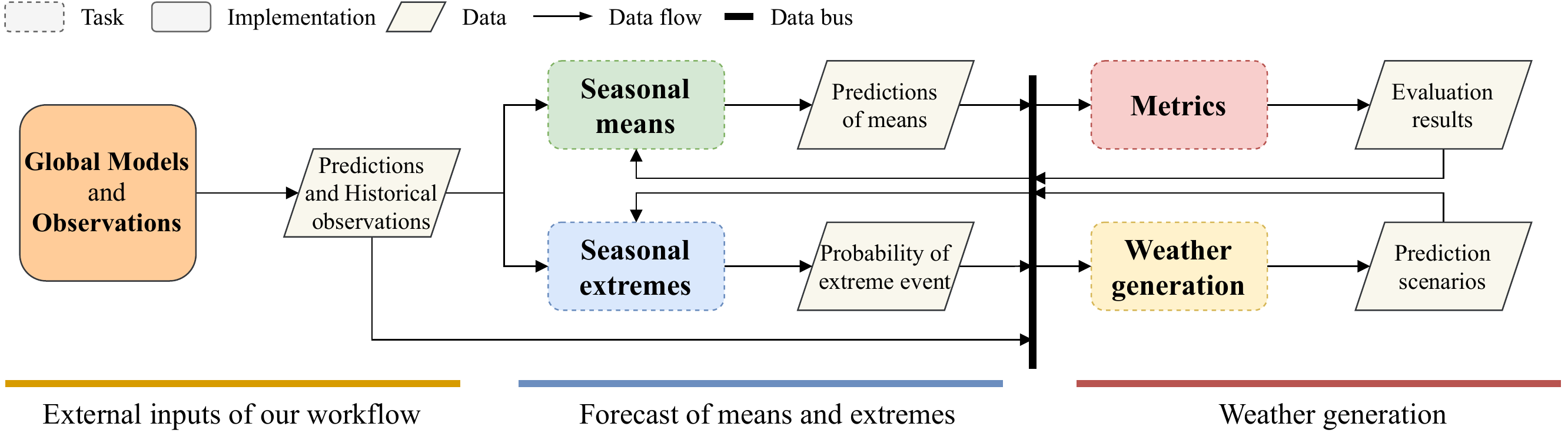}
\caption{Overview of the proposed extreme weather generation framework.}
\label{fig:framework}
\end{figure}

Figure \ref{fig:framework} depicts our proposed framework.
It considers different input data, such as the outputs of global weather models, reanalysis, and observational data.
Such data sources provide rich information about historical records (reanalysis and observational data) and future climate trends (e.g., GCMs).

The \emph{seasonal means} component produces seasonal trend forecasts for different weather variables at different temporal and spatial resolutions.
This component is flexible enough to include models ranging from a simple climatology (i.e., seasonal averages) and statistical models to more sophisticated ML models and physics-based seasonal forecasts (e.g., from SEAS5).
The outputs of the seasonal means component feeds the weather generation component for data calibration purposes.

The \emph{seasonal extremes} component detects weather extremes within historical data and, more importantly, estimates the probability of extreme event occurrence in the future.
As with the \emph{seasonal means}, this component could be instantiated by models ranging from a simple climatology for extreme events (i.e., the observed frequency of the extreme event in historical data) to more sophisticated ML models and well-calibrated physics-based seasonal forecast.
Outputs from this component provide information to the weather generator part about weather extremes definitions and the occurrence probability of future extreme weather events.

The \emph{weather generation} component produces several scenarios for future weather, conditioned or parameterized over the seasonal means and extremes predictions.
We aim to make this component flexible enough to consider hypothetical scenarios not seen in historical data but reasonable under climate change scenarios.
This component could be instantiated using classical stochastic weather generation techniques or through the application of ML techniques such as GANs.

The \emph{metrics} component comprises different types of evaluation metrics.
The idea is to ensure standard ways of evaluating instances of the other components in terms of their ability to predict means, to estimate the probability of extreme events and to generate plausible scenarios.
Examples of metrics that are often used to evaluate probabilistic predictions are the Brier score \cite{Brier1950} and the Continuous Ranked Probability score \cite{Matheson1976}. For weather generation, quantile (Q-Q) plots are commonly used to compare the distribution of the generated data with the distribution of the real data.

\subsection{Seasonal predictions (means and extremes)}


It is common practice (\cite{han2017}) to condition WG parameters on tercile forecasts that specify the probabilities of observing above, near, and below-normal precipitation and temperature or the phase of the El Ni{\~{n}}o Southern Oscillation (ENSO).
For this, probabilistic predictions of the seasonal mean conditions are needed, typically at weekly or monthly aggregation.
There are many examples of ML generating these predictions, such as \cite{yan2020temporal}, where an ensemble of CNNs is created to predict ENSO.

The first step in estimating the probability of extremes is to define what is the event of interest.
Once the event is defined, the most straightforward approach to estimate its probability is to compute an empirical distribution of those events.
For relatively common events and large datasets, this approach converges to the correct distribution.
On the other hand, it may fail to predict rare events, such as the extreme ones.
A more sophisticated strategy is to fit an extreme value distribution to the data. The basic approaches for this (block and threshold methods) \cite{coles2001introduction} assume stationarity.
In the context of climate change, this assumption may not be appropriate. ML methods can be used to learn trends that can improve the estimates for a changing probability of extremes.
Many examples focus on detecting rare events, such as \cite{ding2019modeling}.
The authors use a GRU network to memorize the tail distribution and an attention mechanism to select normal and rare events.
In \cite{laptev2017time}, Bootstrap and Bayesian approaches to estimate uncertainty enable more generalization to an LSTM architecture.

\subsection{Weather generation}


As mentioned before, the weather generation module is responsible for creating realistic weather scenarios given a set of conditions provided by the Seasonal Predictions block.
The idea is to use these conditions as control variables for the weather generation.
For instance, the number of scenarios containing extreme events should respect the probability of such events ($P$).
In this case, the WG will create $P\cdot N$ time series with at least one extreme event and $(1-P)\cdot N$ time series without these events, where $N$ is the number of scenarios.
Additionally, the seasonal forecasts can condition the WG to generate sequences consistent with these future predictions.



In this sense, several WGs were proposed to generate realistic scenarios given different conditions. However, stochastic, statistical, or physical models for weather generation often depend on complex parametrization. We aim to explore ML models to serve as surrogates for weather generation and ease the arduous task of defining such parameters. Different deep learning approaches tackle data synthesis in various problems, and two of the most popular ones are GANs. Following works that use GANs for weather generation \cite{gan1,gan2}, we explore such models' possibilities for a preliminary exercise on weather generation: synthesizing daily precipitation data.



To begin with, we propose  an elementary exercise: to create realistic daily precipitation time series with and without an extreme event. We evaluate how GANs perform compared to the semiparametric multivariate stochastic weather generator proposed by \cite{steinschneider2013semiparametric} and implemented in the \emph{weathergen} library. This baseline weather generator uses a three-state (dry/wet/extreme) first-order Markov chain model per month as an occurrence model, a resampling scheme based on kernel density estimator using weighted data from a KNN algorithm as a precipitation model, and an ARIMA annual precipitation forecaster model for calibration purposes.


\begin{wrapfigure}[18]{r}{0.5\textwidth}
 \centering
 \includegraphics[width=0.48\textwidth]{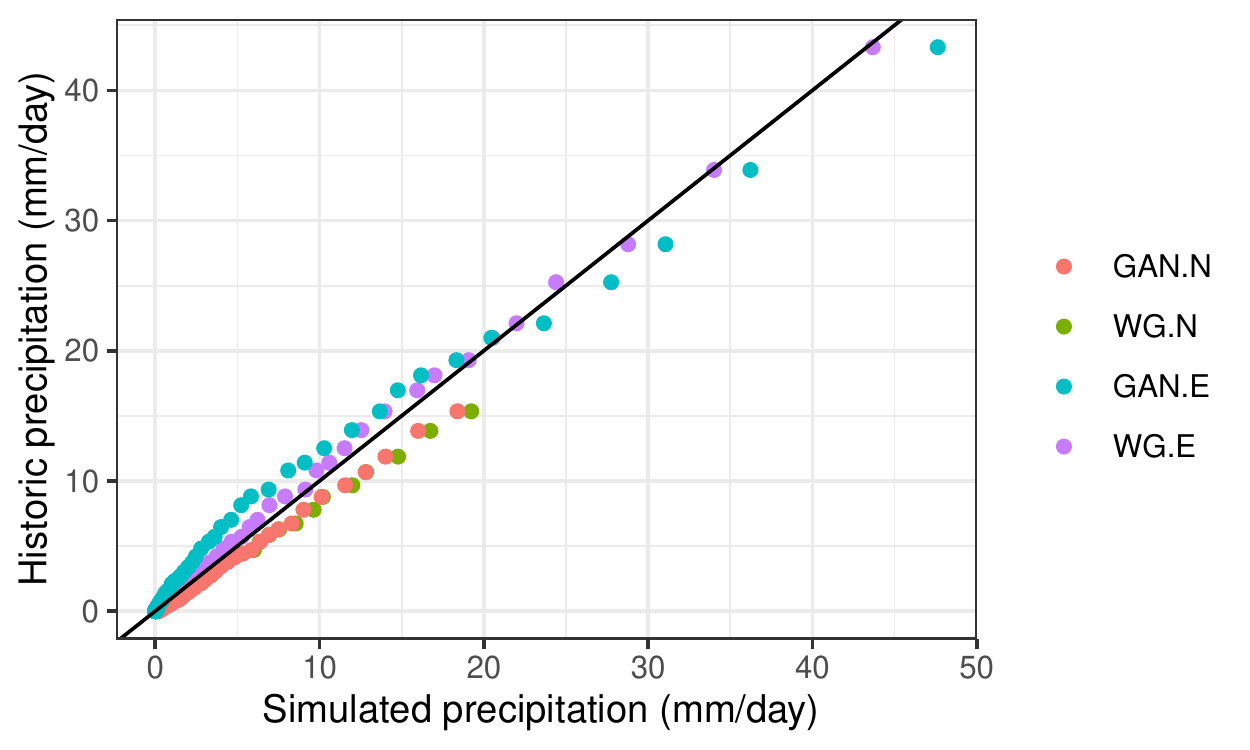}
 \caption{QQ-plot for the simulated values of each model. GAN\_\{E,N\} refers to extreme and normal data using GANs, and WG\_\{E,N\} to \emph{weathergen}. The black line~indicates a perfect match between simulated and historical data.}
 \label{fig:normalQQplot}
\end{wrapfigure}

The GAN architecture was modified to handle one-dimensional signal data. We trained two different models: one using historical data for months with an extreme precipitation event, and the other with regular months. For \emph{weathergen}, we used the same experimental setup.  

The dataset used for training the models comprises a daily 1/8-degree gridded precipitation data from 1 Jan 1949 - 31 Dec 2010 for Boston \cite{maurer2002long}. We defined the monthly extreme threshold as the 95-th percentile of the historical daily precipitation data, which resulted in a value of 18.2 mm/day.

We generated 100 scenarios for months with extreme precipitation events and another 100 for regular months, for each approach. Figure \ref{fig:normalQQplot} shows the QQ-plot of the simulated vs. the historical precipitation time series. We can observe that GANs and \emph{weathergen} perform similarly, indicating that GANs can be used to synthesize realistic daily precipitation data series. While GANs perform slightly better for regular months, \emph{weathergen} performs better than GANs for months with extreme precipitation events.

As \emph{weathergen} uses resampling, it is limited to generate precipitation values that are present in the historical data. Conversely, GANs implicitly learn the distribution of the training data, and a simulated series is any sample from this learned distribution. Moreover, conditional GANs can easily embed control variables and be adapted to work with 2D gridded data. These features can be beneficial in a more challenging experiment that requires gridded multivariate series consistent with climate change projections.

\section{Conclusions}

We presented a modular framework to generate extreme weather event scenarios.
In a climate-changing world, impact and adaptation studies require that such plots be consistent with future climate projections and represent extreme events correctly.
Current WGs usually underestimate extremes and are not flexible concerning conditional variables.
In this sense, ML techniques represent a promising alternative.
By breaking the generation task into simpler subtasks, the proposed workflow allows for a direct application of ML models.
Our initial results applying GANs to generate precipitation series are encouraging considering their potential.
We highlight that our framework is a work in progress and that our next step will contemplate using ML techniques in all components.
Additionally, exploring deep generative models in experiments that consider 2D gridded data and more control variables is necessary to evaluate these models compared to traditional WGs.
The framework will allow such a comparison under the same conditions for both approaches.

\medskip

\small

\bibliography{bibliography}

\end{document}